\def\year{2020}\relax
\documentclass[letterpaper]{article} 

\usepackage[dvipsnames]{xcolor}

\usepackage{aaai20}  
\usepackage{times}  
\usepackage{helvet} 
\usepackage{courier}  
\usepackage[hyphens]{url}  
\usepackage{graphicx} 
\def\UrlFont{\rm}  
\usepackage{graphicx}  
\title{Accelerating and Improving AlphaZero Using Population Based Training}
\author{
Ti-Rong Wu\textsuperscript{\rm 1},  Ting-Han Wei\textsuperscript{\rm 1,3}, I-Chen Wu\textsuperscript{\rm 1,2}\\
\textsuperscript{\rm 1}Department of Computer Science, National Chiao Tung University, Taiwan\\
\textsuperscript{\rm 2}Pervasive Artificial Intelligence Research (PAIR) Labs, Taiwan\\
\textsuperscript{\rm 3}Department of Computing Science, University of Alberta, Edmonton, Canada\\
\{kds285, ting, icwu\}@aigames.nctu.edu.tw 
}
\begin{document}

\maketitle

\begin{abstract}
AlphaZero has been very successful in many games. Unfortunately, it still consumes a huge amount of computing resources, the majority of which is spent in self-play. Hyperparameter tuning exacerbates the training cost since each hyperparameter configuration requires its own time to train one run, during which it will generate its own self-play records. As a result, multiple runs are usually needed for different hyperparameter configurations. This paper proposes using population based training (PBT) to help tune hyperparameters dynamically and improve strength during training time. Another significant advantage is that this method requires a single run only, while incurring a small additional time cost, since the time for generating self-play records remains unchanged though the time for optimization is increased following the AlphaZero training algorithm. In our experiments for 9x9 Go, the PBT method is able to achieve a higher win rate for 9x9 Go than the baselines, each with its own hyperparameter configuration and trained individually. For 19x19 Go, with PBT, we are able to obtain improvements in playing strength. Specifically, the PBT agent can obtain up to 74\% win rate against ELF OpenGo, an open-source state-of-the-art AlphaZero program using a neural network of a comparable capacity. This is compared to a saturated non-PBT agent, which achieves a win rate of 47\% against ELF OpenGo under the same circumstances.

\end{abstract}

\section{Introduction}
Up until recently, games such as chess, Go, and shogi had crucial roles as interesting and challenging measures of development in artificial intelligence research. DeepMind's work, starting with AlphaGo \cite{silver2016mastering}, followed up by AlphaGo Zero \cite{silver2017mastering}, and culminating in AlphaZero \cite{silver2018general}, demonstrated that reinforcement learning can be a powerful tool in solving difficult problems, first with the help of human expert knowledge, then, without any human intervention.

While this family of algorithms were able to deal with the challenge posed by these benchmarks, interest in classical games research remains high. Starting from ELF OpenGo, a reimplementation of the AlphaGo Zero/AlphaZero algorithm, Facebook AI Research is also moving ahead with the more general ELF project \cite{Tian2019ELFOA}, which is aimed at covering a wider range of games for reinforcement learning research. In late August, 2019, DeepMind also announced the OpenSpiel framework, with the goal of incorporating various games with different properties \cite{lanctot2019openspiel}.

Given the continued interest in using games as a reinforcement learning environment, there are still issues that need to be resolved even with a powerful algorithm such as AlphaZero. First, AlphaZero training requires a significant amount of computing resources, at a scale that is prohibitively costly for smaller research teams. As an example, both DeepMind and Facebook AI research use several thousand GPUs to train their Go agents \cite{Tian2019ELFOA}. Recently, Wu \shortcite{Wu2019AcceleratingSL} tested and proposed several techniques that can accelerate AlphaZero training. First, the number of network outputs was increased, including multiple value outputs for different komi\footnote{In Go, komi is the number of points added to the second player to balance the game.} values, and also having new outputs for ownership of board intersections \cite{wu2018multilabeled}. Additionally, Wu also proposed a new method for Monte-Carlo tree search (MCTS) exploration and variation of search parameters, among many other techniques which we will not cover in detail in this paper. While the acceleration factor was reported to be 50, the new techniques introduced a variety of hyperparameters, which were all set to specific values without further explanation.

This leads to the next issue that remains to be resolved for AlphaZero training, namely hyperparameter choice or design. Each hyperparameter configuration requires a significant amount of computing resource commitment before its effects on the trained agent can be observed. As an improvement on manually tuning hyperparameters through experience, Wang et al. \shortcite{wang2019hyper} investigated designing hyperparameter configurations by sweeping each hyperparameter and evaluating their different combinations for 6x6 Othello. By comprehensively listing 12 hyperparameters, then testing each with three different values, they were able to arrive at some intuition on what good hyperparameter design entails in about 36 runs. However, considering its comprehensiveness and the fact that exactly one best hyperparameter configuration is needed ultimately, this method is inefficient in practice. Furthermore, hyperparameter tuning in this case is performed in an offline manner.

In this paper, we propose an online hyperparameter tuning method based on population based training (PBT) \cite{Jaderberg2017PopulationBT}. We can then perform hyperparameter adjustment while the AlphaZero algorithm trains, saving precious computing resources. Another significant advantage of using PBT is that this method requires a single run only while incurring a small additional cost for the optimization and evaluation phases of AlphaZero training.

We test our PBT hyperparameter adjustment method on 9x9 and 19x19 Go, where the PBT method is tested against a baseline of 8 AlphaZero agents, each with its own hyperparameter configuration. We pick two hyperparameters to adjust dynamically, the learning rate and the value loss ratio. Judging by the win rate against the 8 baselines, the PBT method is able to achieve a higher win rate for 9x9 Go. For 19x19 Go, with PBT, we were able to obtain improvements (up to 74\% win rate against ELF OpenGo) in playing strength from a saturated agent (about 47\% win rate against ELF OpenGo). For the two hyperparameters, PBT is shown to be able to decay the learning rate accordingly, while also adjusting the value loss ratio dynamically during training.

\section{Background}
In this section, first, we review the family of AlphaZero-like algorithms. Second, we review the PBT method.

\subsection{AlphaZero-Like Algorithms}

Since AlphaGo successfully defeated the top human Go player in 2016 \cite{silver2016mastering}, DeepMind followed up with the algorithms AlphaGo Zero \cite{silver2017mastering} and AlphaZero \cite{silver2018general}. The main breakthrough for these two successor algorithms is that they do not require any human expert knowledge or input, other than the basic rules of the relevant game. First, we review the AlphaGo Zero algorithm, then we briefly point out the differences between AlphaGo Zero and AlphaZero.

There are three phases during each iteration of AlphaGo Zero training: self-play, optimization, and evaluation. In the self-play phase, the current best network weights (as according to the evaluation phase, which we will discuss shortly) are used to generate self-play records via MCTS. The generated self-play records are then stored in a replay buffer. During the optimization phase, the algorithm samples random positional data from this replay buffer, and uses the sampled batch of data to update the network weights, such that: 
\begin{itemize}
\item the error between the output value $v$ and the sampled data's ground truths $z$ is minimized, and
\item the similarity of the output policy $p$ (consisting of a probability distribution of all moves) and the sampled MCTS search probabilities $\pi$ is maximized.
\end{itemize}
More specifically, during optimization, the parameterized network $\theta$ is updated to minimize the loss 
\begin{equation}\label{formula1}
    L=(z-v)^2 - \pi^{\rm{T}} \log p + 10^{-4} \| \theta\|^2.
\end{equation}
where the last term is the L2 weight regularization. 

For every 1,000 training steps during optimization, the network weights are saved as a checkpoint, each representing an agent that is capable of playing Go with different levels of ability. In the evaluation phase, the new checkpoint is evaluated against the current network. If the checkpoint is superior to the current network, namely, if it surpasses the current network by a win rate of 55\% and above, it replaces the current network. For the next iteration, the best of all networks will be used to generate a new collection of self-play game records.

While AlphaGo Zero was designed to tackle Go, the AlphaZero algorithm \cite{silver2018general} was proposed to generalize to other games, where DeepMind focused on the games of shogi and chess. AlphaZero shares most of the same routines as in AlphaGo Zero, with a few differences. We do not list the details completely, but instead focus on what is one of the most major differences in terms of implementation. In the AlphaZero algorithm, the evaluation phase is removed, and the self-play games are generated instead by the latest network, rather than the superior one, as described above. 

Since the publication of AlphaGo Zero and AlphaZero, there have been numerous open-source projects that have tried to replicate their results. These include Facebook AI Research's ELF OpenGo \cite{Tian2019ELFOA}, the crowd-sourced LeelaZero \cite{Leela-zero}, MiniGo \cite{Lee2019MinigoAC}, written by Google engineers, and KataGo \cite{Wu2019AcceleratingSL}, which was trained using resources from the company Jane Street. KataGo is of particular interest since it can accelerate training by a factor of 50. On the other hand, to achieve this acceleration, a significant number of techniques were used, introducing many new hyperparameters to the overall algorithm.

\subsection{Population Based Training}

Hyperparameter choice is highly critical to whether a neural network based approach to solving a problem succeeds or not. In many cases, the importance of picking the right hyperparameters is made even more apparent since a configuration can only be evaluated after a long period of network training. Despite this, hyperparameter choice often relies on human experience or computationally expensive search algorithms. Recently, population based training (PBT) was proposed to support online training, which adjusts hyperparameters dynamically \cite{Jaderberg2017PopulationBT}. Thereafter, PBT was successfully applied to many problems, most notably on Quake III Arena Capture the Flag \cite{jaderberg2019human}.

Using similar concepts to genetic algorithms, PBT works by training multiple neural networks with initially random hyperparameters. The entire population of networks pool information together to improve the hyperparameters, and concentrates more computational resources on the better-performing individuals. A straight-forward method that implements exploitation involves replacing a lower-performing network with a better-performing network by copying its hyperparameters directly. Similar to mutation in genetic algorithms, there are also several ways of changing current hyperparameter values to explore different configurations.

By constantly performing both exploitation and exploration as the overall network training proceeds, PBT ensures that each individual in the population can perform reasonably well while also guaranteeing that previously unseen configurations are attempted at a specific rate. As a result, PBT can commit computational resources towards more promising hyperparameter configurations, all while training proceeds normally.

We specifically mention four mechanisms here, two for exploitation and two for exploration. For exploitation, the \textit{T-test selection} method involves randomly sampling a target from the population, and evaluating the means of the last 10 episodic rewards for both the network itself and the target. If the target outperforms the network, and it also satisfies Welch's t-test, the network is replaced by the target's parameters and hyperparameters. Next, the \textit{truncation selection} method involves ranking all networks in the population, and replacing the bottom $20\%$ individuals with the top $20\%$. 

For exploration, a method called  \textit{perturb} is used to randomly multiply hyperparameters by 0.8 or 1.2. The \textit{resample} method uses a predefined prior probability distribution and resamples hyperparameters from it. Both exploration mechanisms take place after exploitation, where only the replaced individuals are eligible for exploration.


\section{Our Method}
We now present our method, which incorporates PBT into the AlphaZero algorithm.

In AlphaZero training, the self-play games are typically generated by a single agent with the latest network parameters. However, in our approach, self-play games are generated by a population of $P = 16$ agents. This setting for $P$ is an attempt to follow the findings by Jaderberg et al. \shortcite{Jaderberg2017PopulationBT}, where they tested a variety of values for $P$ for the Atari benchmark and found that $P > 20$ begins to yield an improvement. In our case, $16$ is the closest power of 2 to $20$; for the remainder of this paper, we will use $P=16$.

In each iteration of self-play, we randomly choose 8 pairs from these 16 agents, where each agent uses its own latest network parameters. Namely, each agent will play with exactly one other agent in each iteration of self-play. Compared to the original AlphaZero method, where the games were played by a single agent, who acts both as Black and White, in each game we randomly choose the two agents in a pair so that they alternate as Black and White, to ensure better balancing between the two roles. In fact, diversity can be also increased in this way, since different agents may have their own playing styles. A possible advantage we expect is that the competition among different agents would make it easier to explore the weaknesses of other agents, relative to using a single agent. For simplicity, we avoid performing extra computations in each self-play iteration by generating $1/8$ of the total number of games for each pair. Namely, the total number of self-play games in our method will be equal to the original AlphaZero-like method.

Among the 16 agents, we use PBT to optimize the hyperparameters of the learning rate and the value loss ratio $x$. The value loss ratio $x$ indicates the ratio between the policy loss term and the value loss term. Namely, the parameterized network of agents are updated by minimizing the loss
\begin{equation}\label{formula2}
    L=x(z-v)^2 - \pi^{\rm{T}} \log p + 10^{-4} \| \theta\|^2.
\end{equation}
Compared to the original AlphaZero algorithm, where only one agent was trained in the process, we train all 16 agents in parallel. In addition, in PBT, while different agents use their own sets of hyperparameters to generate self-play games, each agent uses all of the game records generated by all agents for optimization.

Since all 16 agents need to be trained following this method, an additional computation cost for optimization is incurred. More specifically, with $P=16$, the optimization computation cost will be 16 times the cost in the original AlphaZero algorithm. Fortunately, the optimization process tends to be much less costly compared to self-play. To illustrate this, we refer to the paper by ELF OpenGo in which it is stated that 2000 GPUs were used for self-play, while only 8 GPUs were needed for optimization; as a comparison, with half as many simulations, the original AlphaZero training involved 5000 TPUs for self-play and 64 TPUs for optimization \cite{Tian2019ELFOA}. For this reason, the cost of 16 times as much optimization computations only incurs a relatively small additional time cost in the entire training scheme. 

One of the key differences between AlphaZero and its predecessors such as AlphaGo and AlphaGo Zero is that it does not perform evaluation, and instead simply replaces the self-player with the newest optimized agent. In our method, to take account of the multiple number of agents, it is important to evaluate the strengths of all the agents, so that the weaker agents will be replaced by the stronger agents.

It is worth noting that agent strength is not a one dimensional measure and that cycles of strengths are possible, e.g. some agent A may win against B, B against C, and C against A. Ultimately, the goal is to train an agent that has a higher win rate against all other agents. Therefore, during evaluation, we use a round-robin tournament for these 16 agents, where each agent plays 6 games against every other agent, with alternating roles as Black and White for fair comparisons. Namely, during evaluation a total of 90 games are played for each agent, and 720 games in total for one iteration (two agents to a pair for a total of 8 pairs). In our experiments, the total number of self-play games is 5,000 for 9x9 Go and 10,000 for 19x19 Go. Thus, the additional cost for evaluation is also minor when compared with the amount of computation spent on generating self-play records. Altogether, the overhead for the cost of optimization and evaluation in PBT is relatively small. 

To optimize hyperparameters using PBT, we choose the truncation selection method for the exploitation strategy and perturbation for the exploration strategy (see the Background section on PBT) as follows. First, we rank all agents by its win rate in evaluation. If the agent is in the bottom 20\% of the population (namely, if the agent belongs to the bottom 3 when $P = 16$), we simply replace them by copying the weights and hyperparameters from the top 20\% of the population (one-to-one). Next, exploration occurs by perturbing the hyperparameters of the replaced networks by a factor of 1.2 or 0.8. 

In AlphaGo Zero and AlphaZero, the ratio between the loss function for the policy and value networks were set to be equal without any further investigation. Given that PBT allows us to adjust weights online, we wanted to see whether these weights could also be adjusted accordingly. 

\section{Experiments}
In this section, we present our experiment results, performed on 9x9 and 19x19 Go. 

\subsection{Experiments for 9x9 Go}
For the 9x9 Go experiments, the network architecture consists of 3 residual blocks with 64 filters. In our experiment, for each baseline (following the AlphaZero algorithm, trained using the loss function as in Equation (\ref{formula2})), we run a total of 200 iterations, where each iteration contains a self-play phase with 5000 games and an optimization phase. That is, a total of 1,000,000 games are generated for each trained network. Note that the komi of 9x9 Go is 7, leading to the possibility that the outcome may be a draw.

\begin{table}
    \centering
    \begin{tabular}{| c | c | c |}
        \hline 
        Agent ID & Learning rate & Value loss ratio \\
        \hline 
        1 & 2e-2, 2e-3 & 1 \\
        \hline 
        2 & 2e-2, 2e-3 & 1 \\
        \hline 
        3 & 2e-2, 2e-3 & 0.5 \\
        \hline 
        4 & 2e-2, 2e-3 & 2 \\
        \hline 
        5 & 2e-2 & 1 \\
        \hline 
        6 & 2e-2 & 1 \\
        \hline 
        7 & 2e-2 & 0.5 \\
        \hline 
        8 & 2e-2 & 2 \\
        \hline 
        PBT & 2e-2 & 1 \\
        \hline 
    \end{tabular}
    \caption{The hyperparameters for the 8 baselines, all of which are based on AlphaZero training, and the initial values for the PBT version.}
    \label{tab:8 P+V Baseline}
\end{table}

We train a PBT with $P = 16$ agents, and also 8 AlphaZero versions individually as baselines. Although the baseline versions only consist of 8 agents, which is less than PBT which has a total of 16 agents, the computation cost of training the baseline versions is almost 8 times that of PBT. Following the example set by AlphaGo Zero \cite{silver2017mastering} and AlphaZero \cite{silver2018general}, we randomly initialized the network parameters and set different learning rates and value loss ratios\footnote{In fact, we trained Agents 1, 2, 5 and 6 first, both sets of which follow the AlphaZero hyperparameters, since they are highly regarded in the community. That is, Agents 1 and 2 follow the same settings, while 5 and 6 follow similar settings, but with a constant learning rate. We then halved/doubled the value loss ratio to explore different settings.}, as detailed in Table \ref{tab:8 P+V Baseline}. Note that in Table \ref{tab:8 P+V Baseline}, the learning rate changes after 100 iterations, separated by a comma. In addition, the listed values for PBT are only the initial values, where the hyperparameters will be changed dynamically during training.

\begin{figure}[t]
    \centering
    \includegraphics[width=\columnwidth]{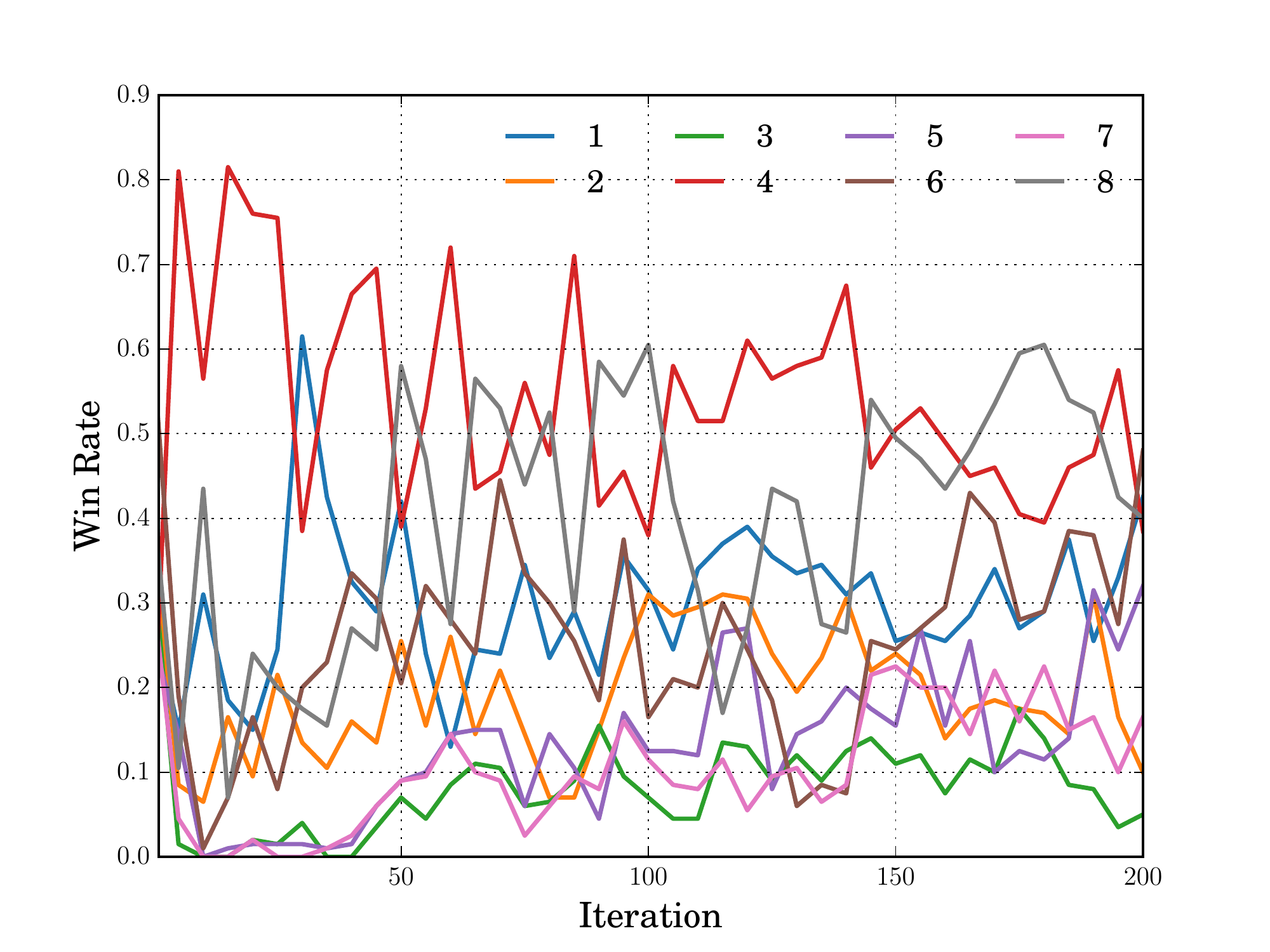}
    \caption{The round-robin results of 8 baseline versions. The win rates shown here are the minimum win rates against the other 7 baselines.}
    \label{fig:8 P+V Baseline Min win rate}
\end{figure}

\begin{figure}[t]
    \centering
    \includegraphics[width=\columnwidth]{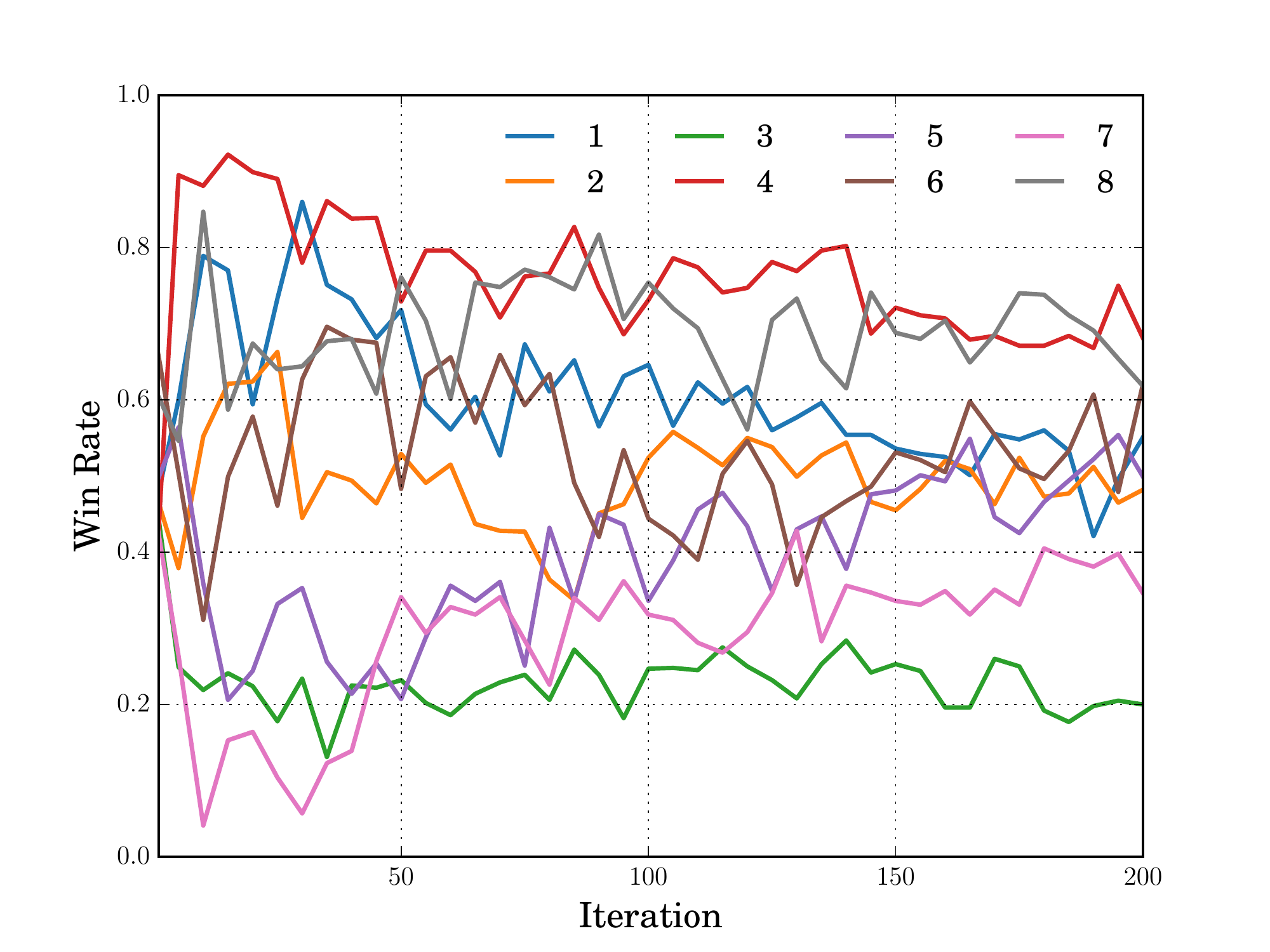}
    \caption{The round-robin results of the 8 baselines versions, where the average win rate is shown.}
    \label{fig:8 P+V Baseline Avg win rate}
\end{figure}

To analyze the performance of the 8 baseline versions, we made a round-robin tournament. Every five iterations, the 8 baselines each play against the other 7 baselines, where each match-up consists of 100 games (for a total of 700 game per baseline). Figure \ref{fig:8 P+V Baseline Min win rate} shows the minimum win rate for each agent; namely, after each agent plays against the other 7 agents, we simply depict the worst win rate in the figure. Figure \ref{fig:8 P+V Baseline Avg win rate} shows the average win rate for each agent since the minimum win rate does not give the full picture of the agent's overall strength.

From these two figures, the result shows that overall agents 4 and 8 perform better than the others. Agent 4 especially stands out, as it performs around 60\% minimum win rate and around 80\% win rate during iterations 100 to 150. From this result, according to the hyperparameters, we observe that the larger value loss ratio of 2 might accelerate and improve training in a total of 200 iterations, while the smaller value loss ratios such as 0.5 perform the worst in this experiment, as is the case with agents 3 and 7. The learning rate schedules are also important, since agent 8 has the same value loss ratio with agent 4, but it performs worse than agent 4 after 100 iterations, which is likely the result of its static learning rate. Interestingly, no single agent can dominate all other agents; for certain iterations (say, around iteration 160 in Figure \ref{fig:8 P+V Baseline Min win rate}), we can even see how every agent's minimum win rate is below 50\%. This is a fitting example to illustrate circular strengths (i.e. how the strength of agents is not a one-dimensional scale) that we mentioned in regards to the evaluation phase in the Method section.

\begin{figure}[t]
    \centering
    \includegraphics[width=\columnwidth]{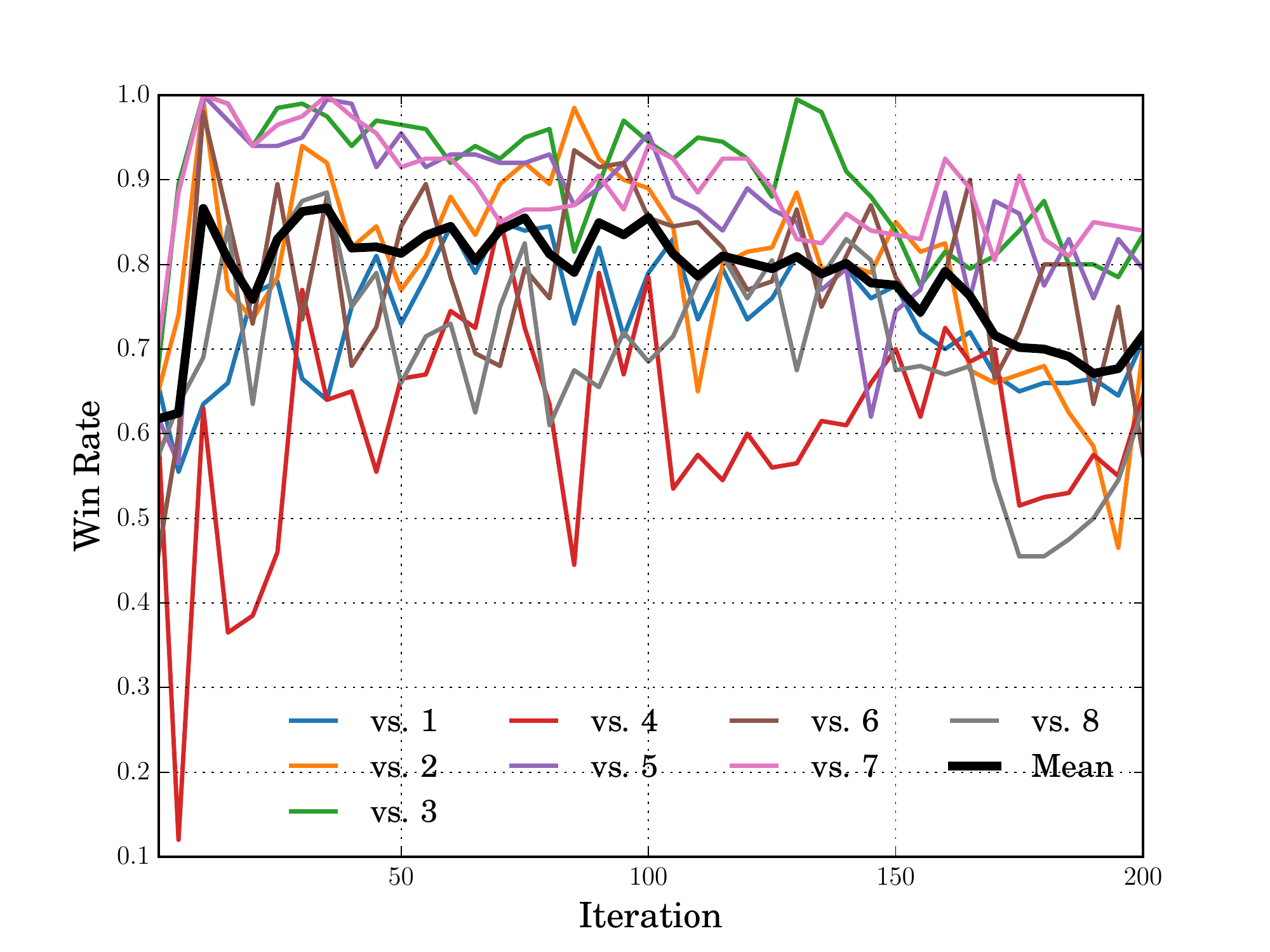}
    \caption{The win rate for PBT against all 8 baselines. Note that each line refers to PBT's win rate; a value of 0.5 or higher indicates that PBT is stronger.}
    \label{fig:P+V win rate}
\end{figure}

Next, we analyze the performance of training by the PBT method. To determine its relative strength to the baselines, for each iteration, we simply use the top agent (out of all 16 agents in the population) that was chosen in that iteration's evaluation phase. The chosen agent then plays against all 8 baselines, where the results are shown in Figure \ref{fig:P+V win rate}. Note that in this figure the average values are depicted as the bolded black line. From the figure, we can see that with the exception of a few iterations where PBT performs slightly worse than one or two baselines (e.g. 45.5\% win rate against agent 8 at iteration 175), its win rate exceeds 50\% in most iterations. Again, we want to stress that a major benefit of using PBT is that it can outperform the 8 baselines without having to generate more self-play games, and consequently saving computation resources by only having to train once.

\begin{figure}[t]
    \centering
    \includegraphics[width=\columnwidth]{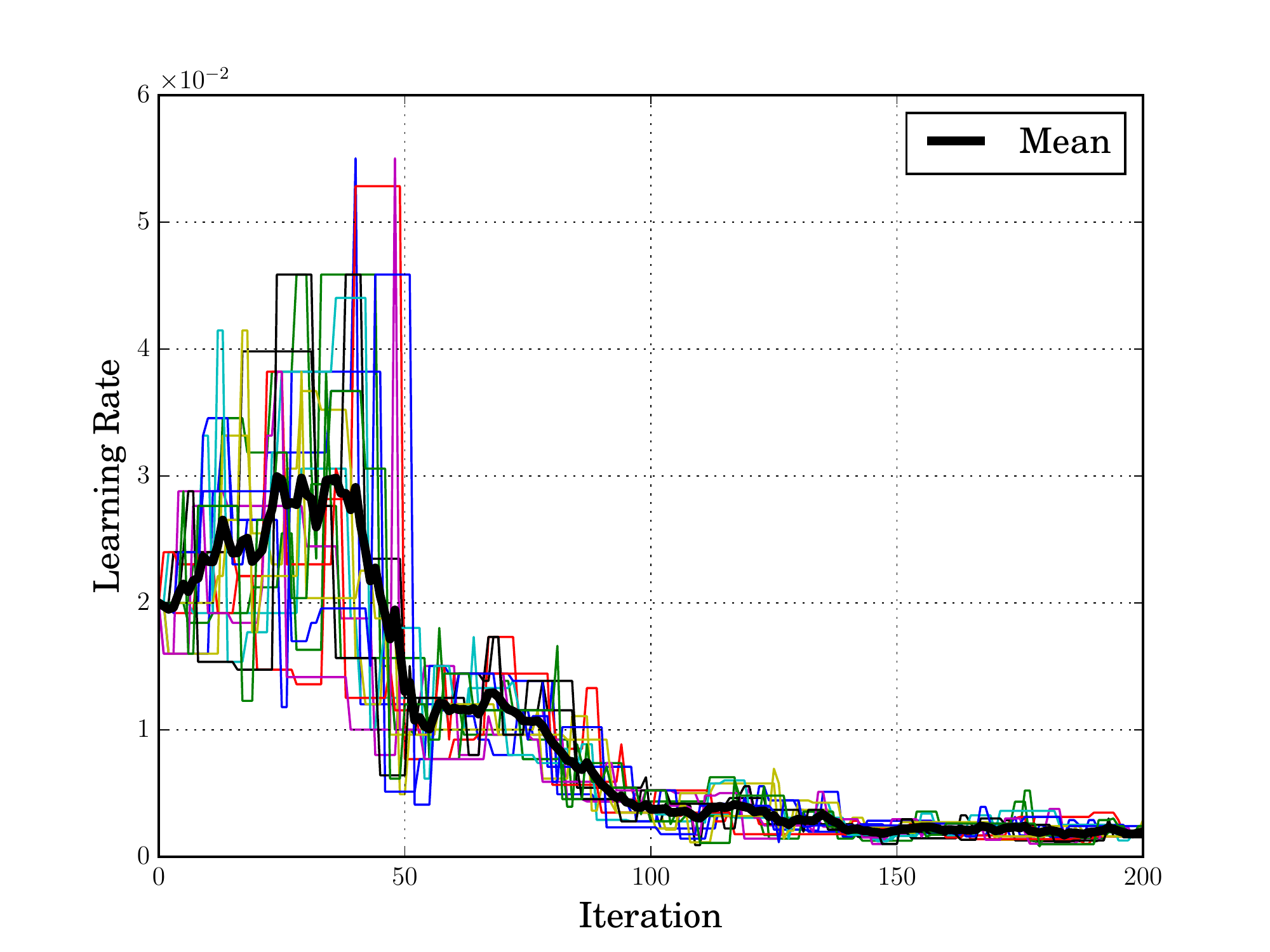}
    \caption{The learning rate curve during training. The colored lines depict the learning rate settings for different agents in the population. The solid black line depicts the mean of all agent learning rates.}
    \label{fig:P+V learning rate}
\end{figure}

\begin{figure}[t]
    \centering
    \includegraphics[width=\columnwidth]{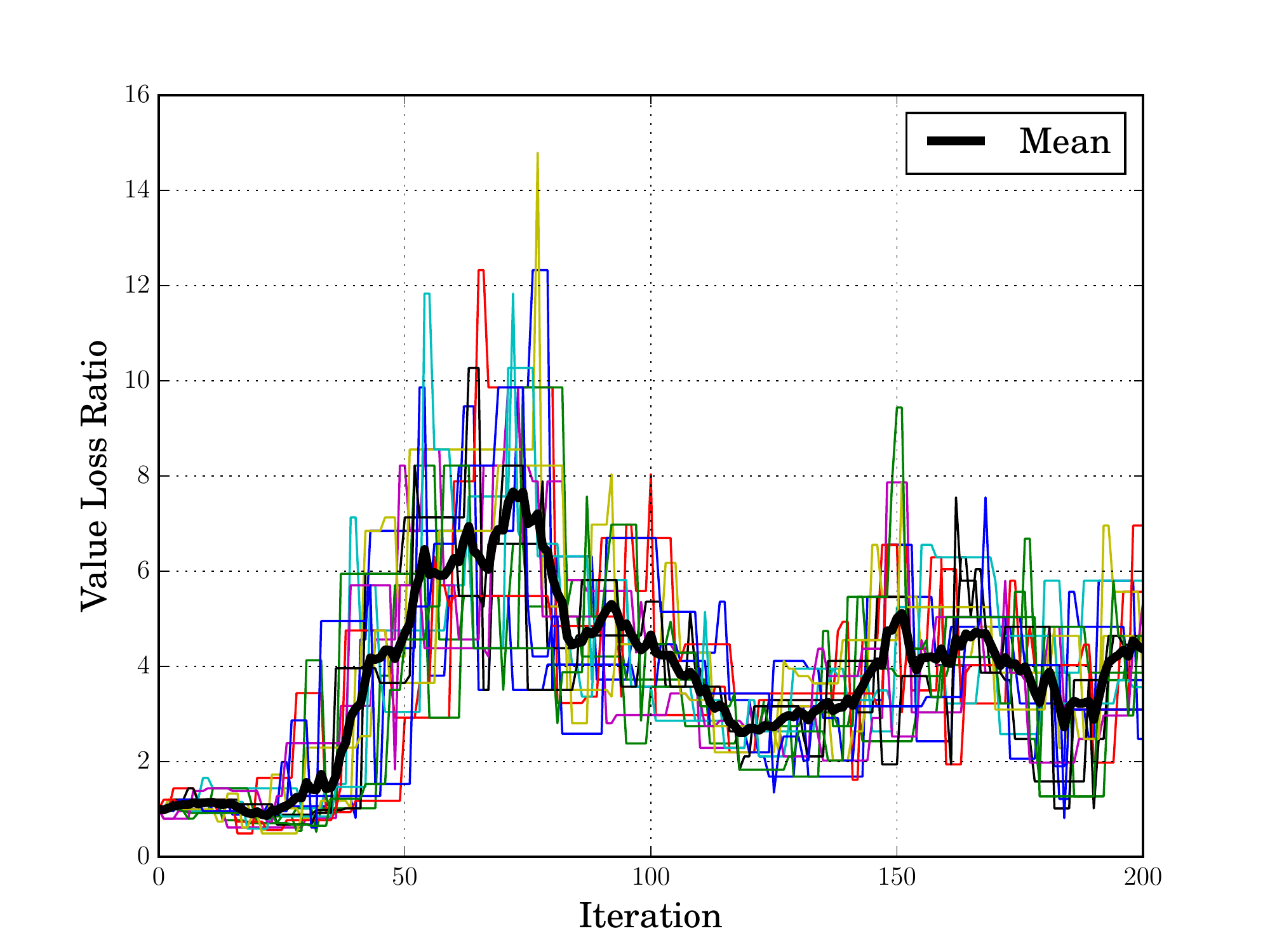}
    \caption{The value loss ratio curve during training. The colored lines depict the settings for different agents in the population. The solid black line depicts the mean.}
    \label{fig:P+V value loss}
\end{figure}

Figure \ref{fig:P+V learning rate} and Figure \ref{fig:P+V value loss} show the trend of the hyperparameters adjusted by PBT. First, in Figure \ref{fig:P+V learning rate}, the average learning rate starts from about 0.02, increase significantly (say, to a maximum of 0.03 at iteration 34), then drops rapidly to 0.015 at around iteration 50 because of the exploit mechanism. The average of the learning rate decrease gradually and reaches 0.003 at iteration 100. After 100 iterations, the learning rate stays at a value around 0.002 to 0.003, which continues up to iteration 200. Interestingly, the schedule seems to be similar to the hand-tuned schedule, as is the case for the baselines. This shows that PBT can adjust learning rates automatically.

Second, in Figure \ref{fig:P+V value loss}, the value loss ratio starts from 1, increases to an average of 7.5 around iteration 70, and then decreases to an average of 2 around iteration 120. Beyond that point, the value loss ratio fluctuates around 2 to 4. In this experiment, we can see that a sufficiently large value loss ratio is better for training, which corroborates our experience with the baseline experiments. While PBT is able to find the best value loss ratio to be around 2 to 4, without PBT we would need more than 8 experiments to figure this out. However, according to the experiments by Tian et al. \shortcite{Tian2019ELFOA} (referred to in the paper as the "dominating value gradients" experiments), if the value loss ratio is too large (say, a value of 361 for 19x19 Go), the agent strength will hit a limitation. Fortunately, PBT offers a dynamic adjustment mechanism that can decrease the value loss ratio when it becomes a limiting factor on performance.

Another interesting observation is that the learning rate and the value loss ratio seem to complement each other in the early stages of training. While the learning rate increases around iterations 0 to 30, the value loss ratio maintains the same value. During iterations 30 to 60, the learning rate decreases, but the value loss ratio increases. Our conjecture is that PBT focuses on the hyperparameters that have a higher impact on performance. Consequently, the learning rate changes more significantly in the beginning stages of training, but then the algorithm "shifts focus" and tunes the value loss ratio once the learning rate stabilizes.

\begin{figure}[t]
    \centering
    \includegraphics[width=\columnwidth]{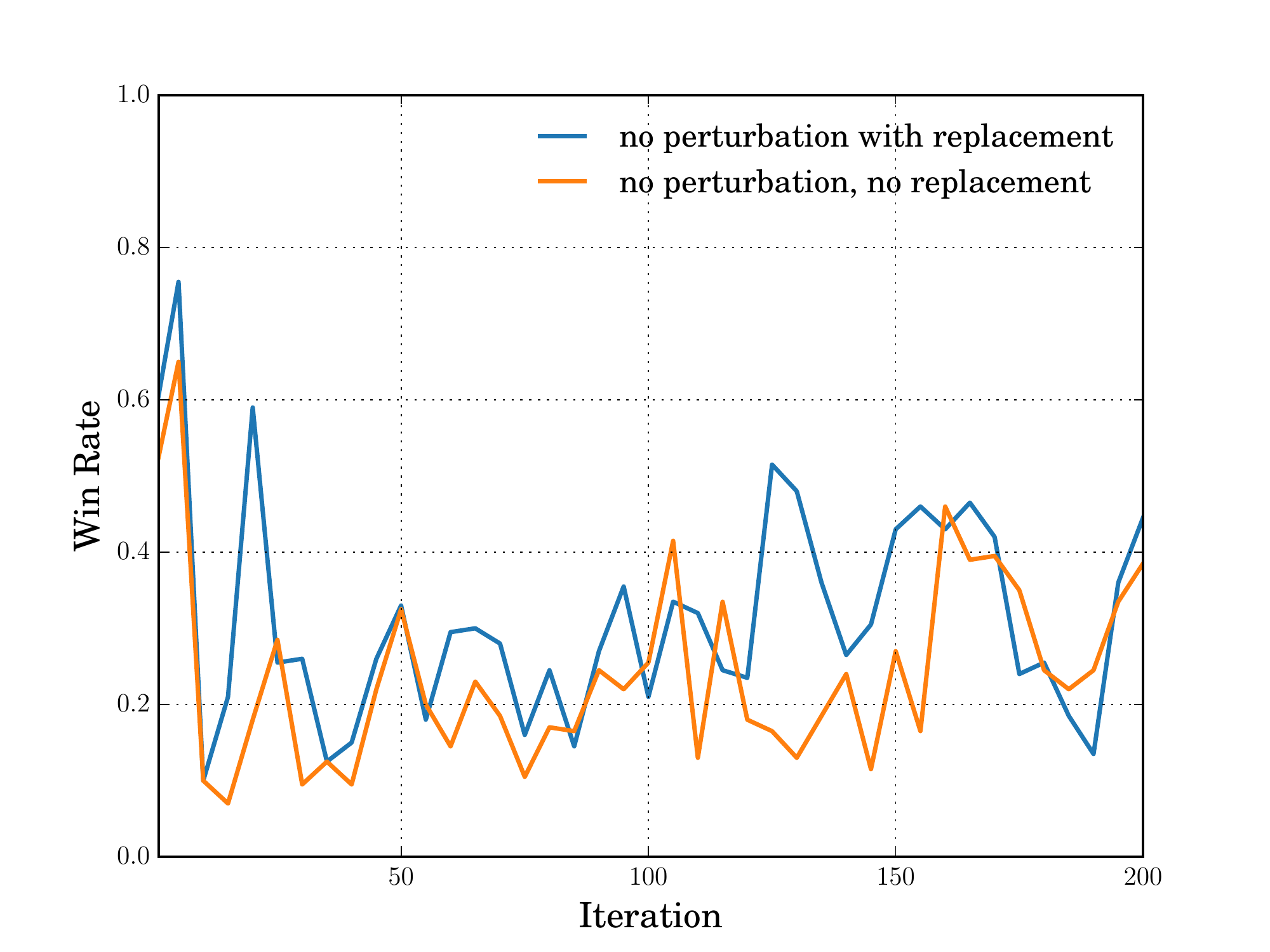}
    \caption{The win rates for the two ablation methods against the PBT method.}
    \label{fig:ablation}
\end{figure}

In addition, we performed two ablation experiments to further analyze the benefit of the PBT method: (1) no perturbation with replacement; (2) no perturbation, no replacement. Since these two experiments are trained without perturbation, we initialized the hyperparameters of the 16 agents for diversity and fairness such that each setting entry in Table \ref{tab:8 P+V Baseline} will have 2 agents. Figure \ref{fig:ablation} shows the win rates for the two ablation methods against the PBT method. Between the two non-perturbed cases, the one with replacement performs slightly better than the without replacement. Although all agents used different initial hyperparameters, in the case with replacement, the agents soon converged to the same hyperparameters. More specifically, the value loss ratio converged to 2 after only 15 iterations. As a result, the training is equivalent to multi-agent training without diversity. Next, for the case with no perturbation and no replacement, the diversity of agents remains high throughout training. However, a full sweep for optimization would involve suboptimal agents that will never be replaced. Since each agent will train using the collection of self-play records generated by all agents, optimization may be negatively impacted by low-quality game records. Conclusively, these ablation experiments show that: (1) PBT performs well not only due to the diversity of agents (as is the case with no perturbation and no replacement); and that (2) PBT with perturbation and replacement can maintain diversity as well as strength, as shown in Figure \ref{fig:ablation}.

Moreover, we performed additional experiments on multi-labelled value networks (MLVN) \cite{wu2018multilabeled} using PBT. MLVN is a simple technique that was also included in KataGo, where it was referred to as the score belief network \cite{Wu2019AcceleratingSL}. The experiment results are similar to those for the single value output network and it shows that the PBT method can generalize for different network architectures.

\subsection{Experiments for 19x19 Go}
Next, we apply our method to our 19x19 Go program CGI \cite{wu2018multilabeled}. We made some changes between 9x9 and 19x19 training as follows. The network is expanded to consist of 20 residual blocks with 256 filters, the same size as ELF OpenGo v2. The output of the network consists of a policy and 31 value outputs, each corresponding to a different komi ranged from -7.5 to 22.5, centered at 7.5 komi. In each iteration, 10,000 games are generated via self-play.

In this experiment, we first train a network following the AlphaGo Zero algorithm, with learning rate decreasing from 0.01 to 0.0001 (following the scheduling by AlphaGo Zero \cite{silver2017mastering}), and finally fine-tuning with a learning rate of 0.00005 after the network is saturated at 0.0001. Saturation in this context refers to the situation where millions of games of additional training no longer leads to improvement on win rates against ELF OpenGo v2. 
Training on the chosen network stopped at saturation with a learning rate of 0.00005.

Next, with the saturated trained network, we apply the PBT method to try and improve the network's performance beyond saturation. For the first part of the PBT training, we increase the learning rate to 0.0001 (from the previous saturated value of 0.00005) for the following reason. Since 0.00005 is already lower than all the published learning rates from current AlphaZero-related articles, we wish to leave some room so that perturbation of the learning rate will not lead to an unreasonably small learning rate. In addition, the ratio between the losses for policy and the multiple value outputs is set to be 0.2 (the same as the baseline version). In our 19x19 training, we start all 16 agents from the same network weights, but with perturbation before training starts to increase the population diversity.

\begin{figure}[th!]
    \centering
    \includegraphics[width=\columnwidth]{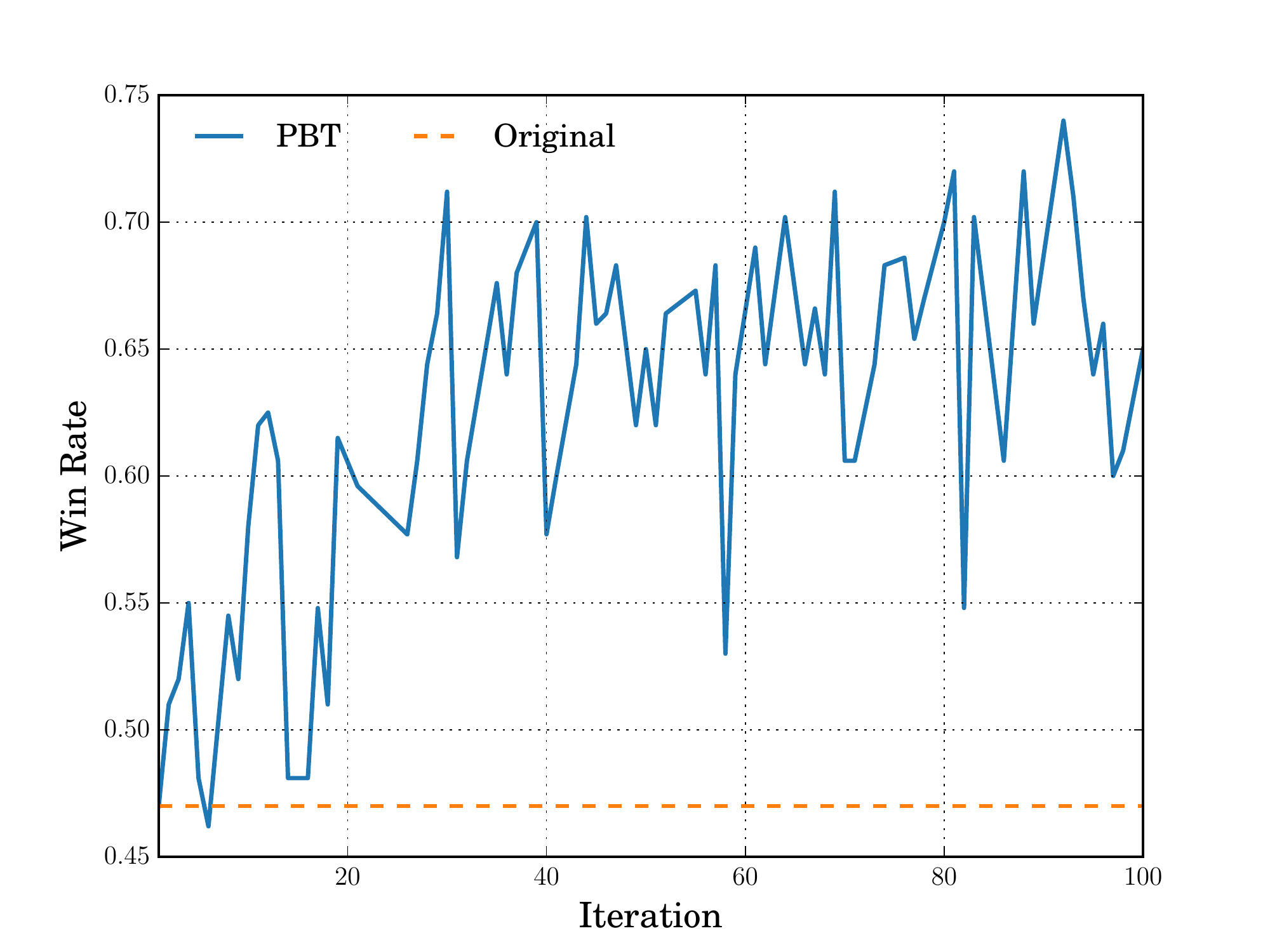}
    \caption{The training curve for the win rate against ELF OpenGo. }
    \label{fig:19x19 win rate}
\end{figure}

\begin{figure}[th!]
    \centering
    \includegraphics[width=\columnwidth]{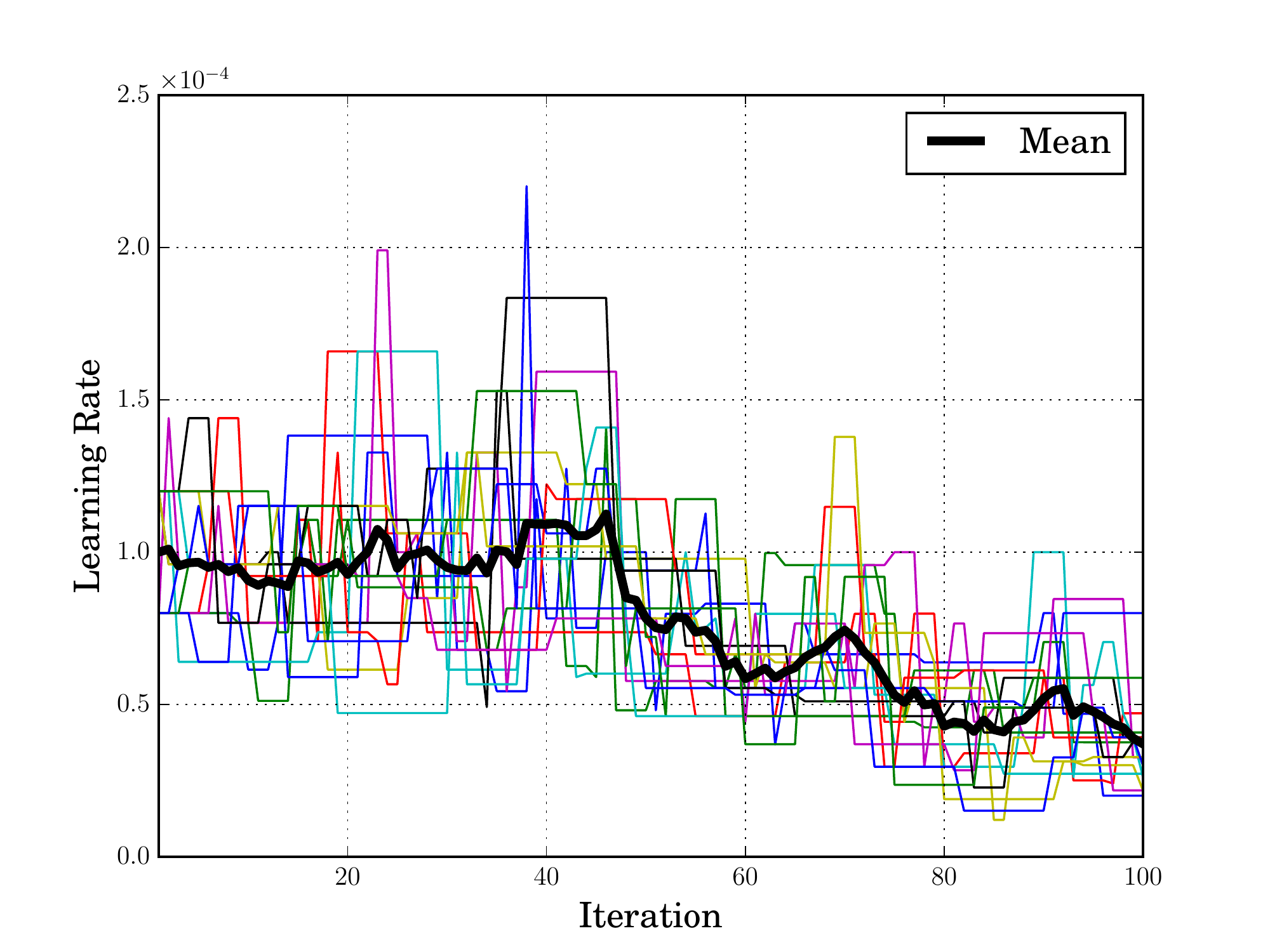}
    \caption{The learning rate curve for 19x19 Go.}
    \label{fig:19x19 learning rate}
\end{figure}

\begin{figure}[th!]
    \centering
    \includegraphics[width=\columnwidth]{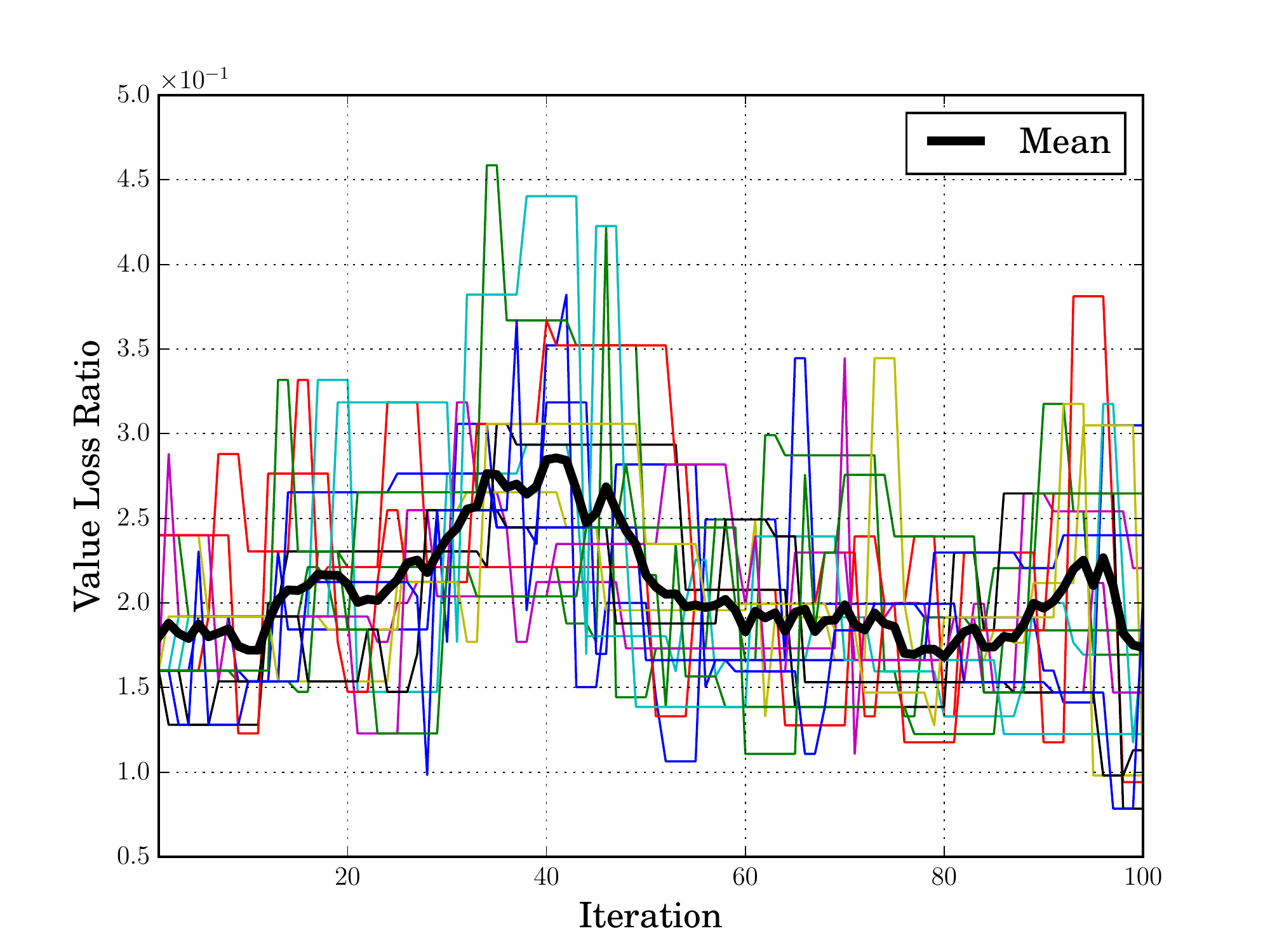}
    \caption{The value loss ratio curve for 19x19 Go.}
    \label{fig:19x19 value loss}
\end{figure}

Figure \ref{fig:19x19 win rate} shows the training curve. The results show that with a network (with the same size as ELF OpenGo v2), after 30 iterations the win rate is 71.2\% against it, while the version without PBT showed no improvement after an additional 100 iterations of training. Moreover, after 92 iterations the PBT version reaches 74.0\% win rate against ELF OpenGo v2.

Figure \ref{fig:19x19 learning rate} and Figure \ref{fig:19x19 value loss} shows the hyperparameters that are adjusted by PBT in 19x19 Go. First, in Figure \ref{fig:19x19 learning rate}, the average values of the learning rate start from 0.0001, and gradually drops to 0.00005 after 80 iterations. Although a few agents use a larger learning rate of, say, 0.0002 in iteration 38, the result shows that the network is at saturation with the larger learning rate, and that improvement can only be achieved by a smaller value. 

Second, the average movement of value loss ratio is shown in Figure \ref{fig:19x19 value loss}. The value loss ratio increases gradually and reaches about 0.3 at iteration 40, where some agents use even larger ratios such as 0.45 for a while. However, after iteration 50, the average movement of the value loss weight drops to 0.2, and only slightly increases after iteration 85. It is interesting to see that the value loss weight changes during different training stages, and that the results are similar to 9x9 Go in the previous subsection. In AlphaZero training, the policy and value complement each other; a stronger policy tends to generate strong values in self-play, and stronger values will generate a stronger policy in MCTS. Thus, in our opinion, it is reasonable that the policy and the value loss weight will fluctuate during training since the agent should pay more attention to the policy or the value at different training stages. PBT therefore offers a dynamic approach to adjust these hyperparameters.

\section{Conclusion}

AlphaZero is a powerful reinforcement learning algorithm that is able to train super-human level agents for many different games without requiring human expert knowledge. However, AlphaZero algorithms often involve many hyperparameters, especially if we wish to accelerate the overall training process.

This paper shows that PBT is a promising method to help tune the hyperparameters, and in turn can be used to improve AlphaZero-like algorithms. Using PBT, by simply adjusting two hyperparameters, the learning rate and the ratio of value to policy loss, we were able to train a 19x19 Go program with a win rate of 74.0\% against Facebook's ELF OpenGo v2, a state-of-the-art open-source 19x19 Go program with 20 residual blocks. To our knowledge, our program, which incorporates PBT into AlphaZero, is state-of-the-art in playing strength among neural networks of a comparable capacity. Other open-source AlphaZero-like reimplementations including MiniGo and KataGo are all reported to be of similar playing strength as ELF OpenGo v2 with 20 residual blocks. This implies that PBT plays a crucial role in penetrating the performance ceiling of state-of-the-art 19x19 Go programs.

We also greatly reduce computing resource usage by leveraging PBT while reaching state-of-the-art performance. Since AlphaZero was first published, many open-source computer Go projects have attempted to reproduce and improve upon it. Unfortunately, training with AlphaZero consumes a tremendous amount of computing resources. As reported by \cite{Tian2019ELFOA}, it requires 2000 GPUs (V100s) over 9 days for training a single run. With different hyperparameters settings (like the hyperparameter sweep as proposed in \cite{wang2019hyper}), this requirement in computing resources will increase accordingly. In this paper, our method can reap the benfits of having a wide collection of hyperparameters, while only requiring a single run with a small extra overhead.

This paper is a simple demonstration that shows how PBT can adjust hyperparameters on-line. A future direction for investigation is applying PBT to a more comprehensive list of hyperparameters. This would include hyperparameters such as the loss ratio for auxiliary outputs (intersection ownership \cite{wu2018multilabeled,Wu2019AcceleratingSL}, long-term prediction \cite{tian2015better}, etc.), the constant $c$ in PUCT \cite{rosin2011multi}, virtual loss during self-play, optimization batch size, L2 regularization term weighting, among others. While it is true that with more hyperparameters, the search space becomes larger, PBT traverses this search space in a more informed manner. It is worth emphasizing that the additional computation cost for using PBT is dependent on the population size, and not directly related to the number of hyperparameters that PBT aims to adjust.

\section{Acknowledgments}

This research is partially supported by the Ministry of Science and Technology (MOST) of Taiwan under Grant Number MOST 107-2634-F-009-011 and MOST 108-2634-F-009-011 through Pervasive Artificial Intelligence Research (PAIR) Labs. The computing resource is partially supported by National Center for High-performance Computing (NCHC) of Taiwan. The authors would like to thank anonymous reviewers for their valuable comments.


\clearpage
\maketitle

\section{Supplementary Materials: Experiments for 9x9 Go Using Multi-Labelled Value Networks}
We present an alternate set of experiments for 9x9 Go, where instead of the single value output, we now use the multi-labelled value network (MLVN) \cite{wu2018multilabeled}, which was also used to improve the performance in KataGo \cite{Wu2019AcceleratingSL} (referred to there as the score belief network). These experiments were performed to see how well the PBT method generalizes for different network architectures.

The MLVN simply changes the single value output to multiple value outputs, each of which corresponds to a different komi in the game of Go. We use a total of 11 komi settings for the value output, from komi 2 to 12 (centered at komi 7). Since it becomes uncertain about the optimal value loss ratio for the multiple outputs, it is interesting to see how PBT will adjust the ratio, when compared with that with single value only.

Similar to the previous experiment, we train using PBT with 16 agents, along with 8 AlphaZero baselines. All versions use the MLVN technique, where the initial hyperparamter values are shown in Table \ref{tab:8 P+ML Baseline}. For agents 1, 5, and the PBT version, we set the value loss ratio as 1 (the default setting by Wu et al. \shortcite{wu2018multilabeled}). For other baselines, the ratio is set to 0.1, 0.2, and 0.5 since with MLVN the total value output is 11 times the original single value output, so we attempt smaller value loss ratios.

\begin{figure}[hbtp!]
    \centering
    \includegraphics[width=\columnwidth]{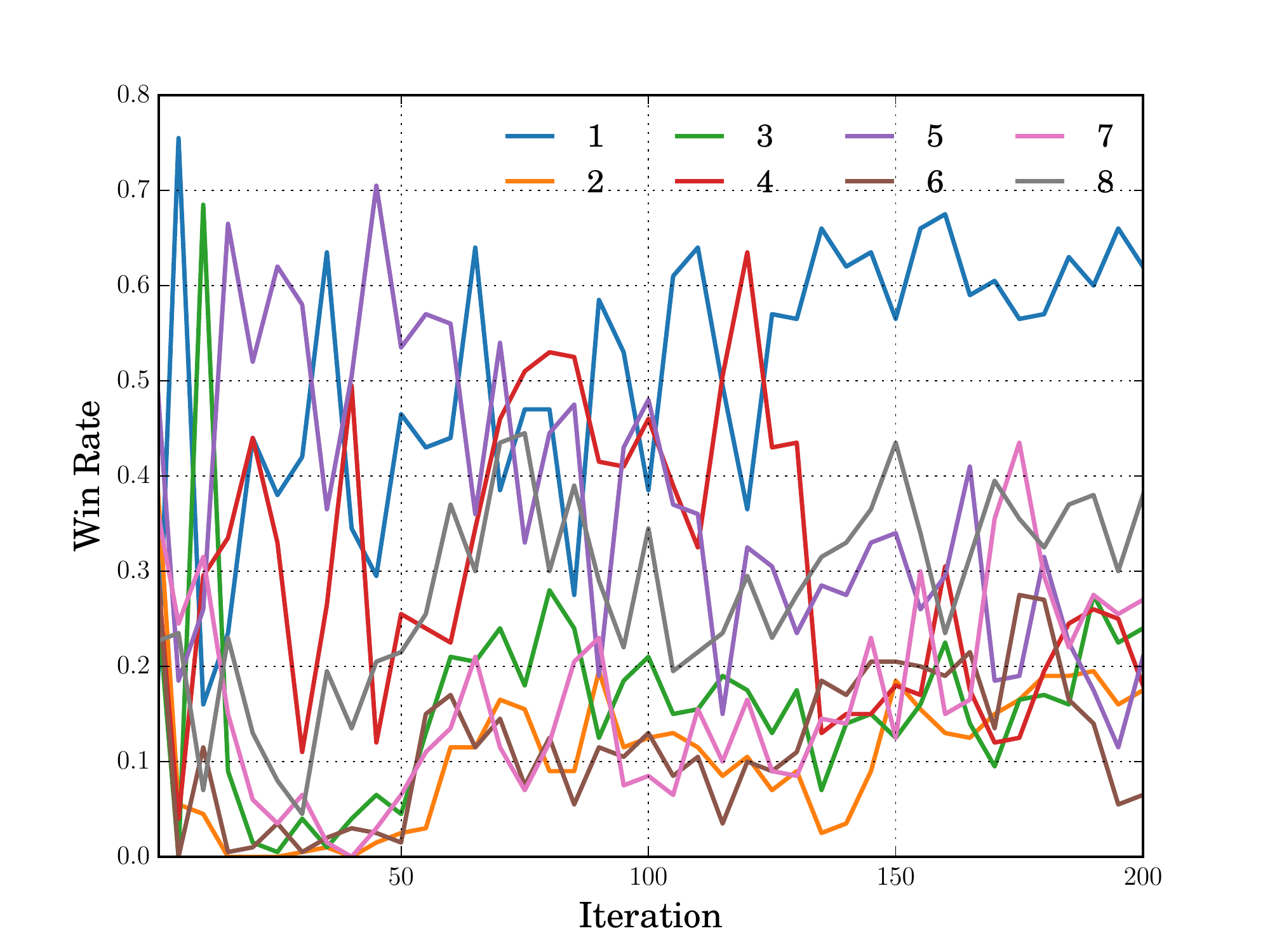}
    \caption{The round-robin results of 8 baseline versions with ML value networks. The win rates shown are the minimum win rates.}
    \label{fig:8 P+ML Baseline Min win rate}
\end{figure}

\begin{figure}[hbtp!]
    \centering
    \includegraphics[width=\columnwidth]{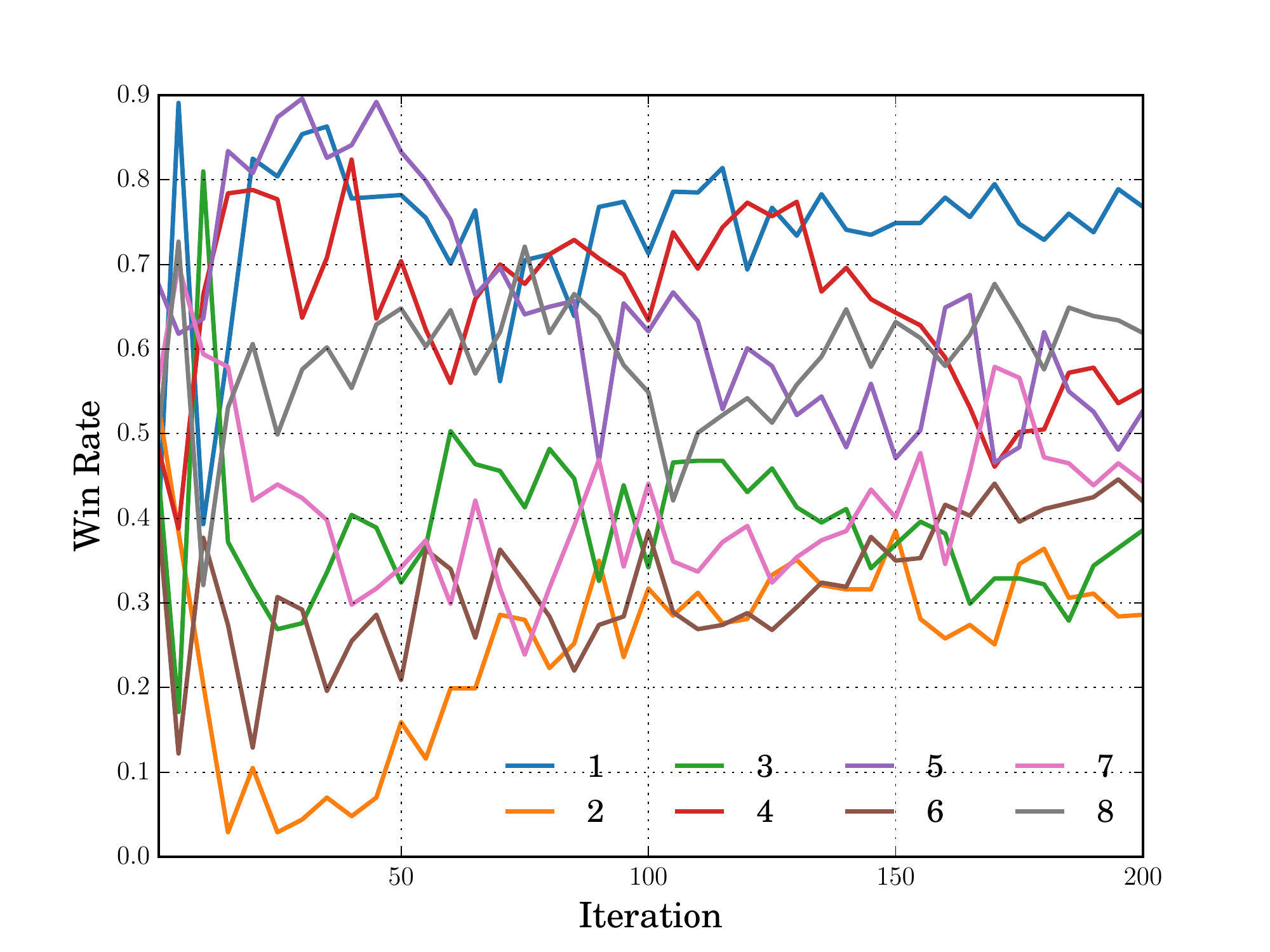}
    \caption{The round-robin results of 8 baseline versions with ML value networks, where the average win rates are shown.}
    \label{fig:8 P+ML Baseline Avg win rate}
\end{figure}

We performed round-robin tournaments for the baselines, similar to the previous experiment. Figure \ref{fig:8 P+ML Baseline Min win rate} and Figure \ref{fig:8 P+ML Baseline Avg win rate} show the minimum win rate and the average win rate for each agent. Surprisingly, agent 1, with the highest value loss ratio, performs much better than all other agents, with almost over 60\% minimum win rate and over 70\% average win rate compared to the other agents. On the other hand, agent 2 and 6, which have smaller value loss ratios, perform worse against all other agents. This result also corroborates the previous baseline experiment: larger value loss ratios with appropriate learning rate decay often leads to good training results.

\begin{table}[hbtp!]
    \centering
    \begin{tabular}{| c | c | c |}
        \hline 
        Agent ID & Learning rate & Value loss ratio \\
        \hline 
        1 & 2e-2, 2e-3 & 1 \\
        \hline 
        2 & 2e-2, 2e-3 & 0.1 \\
        \hline 
        3 & 2e-2, 2e-3 & 0.2 \\
        \hline 
        4 & 2e-2, 2e-3 & 0.5 \\
        \hline 
        5 & 2e-2 & 1 \\
        \hline 
        6 & 2e-2 & 0.1 \\
        \hline 
        7 & 2e-2 & 0.2 \\
        \hline 
        8 & 2e-2 & 0.5 \\
        \hline 
        PBT & 2e-2 & 1 \\
        \hline 
    \end{tabular}
    \caption{The hyperparameters for the 8 baselines and the initial values for the PBT version, all of which are based on AlphaZero training with an ML value network.}
    \label{tab:8 P+ML Baseline}
\end{table}

\begin{figure}[hbtp!]
    \centering
    \includegraphics[width=\columnwidth]{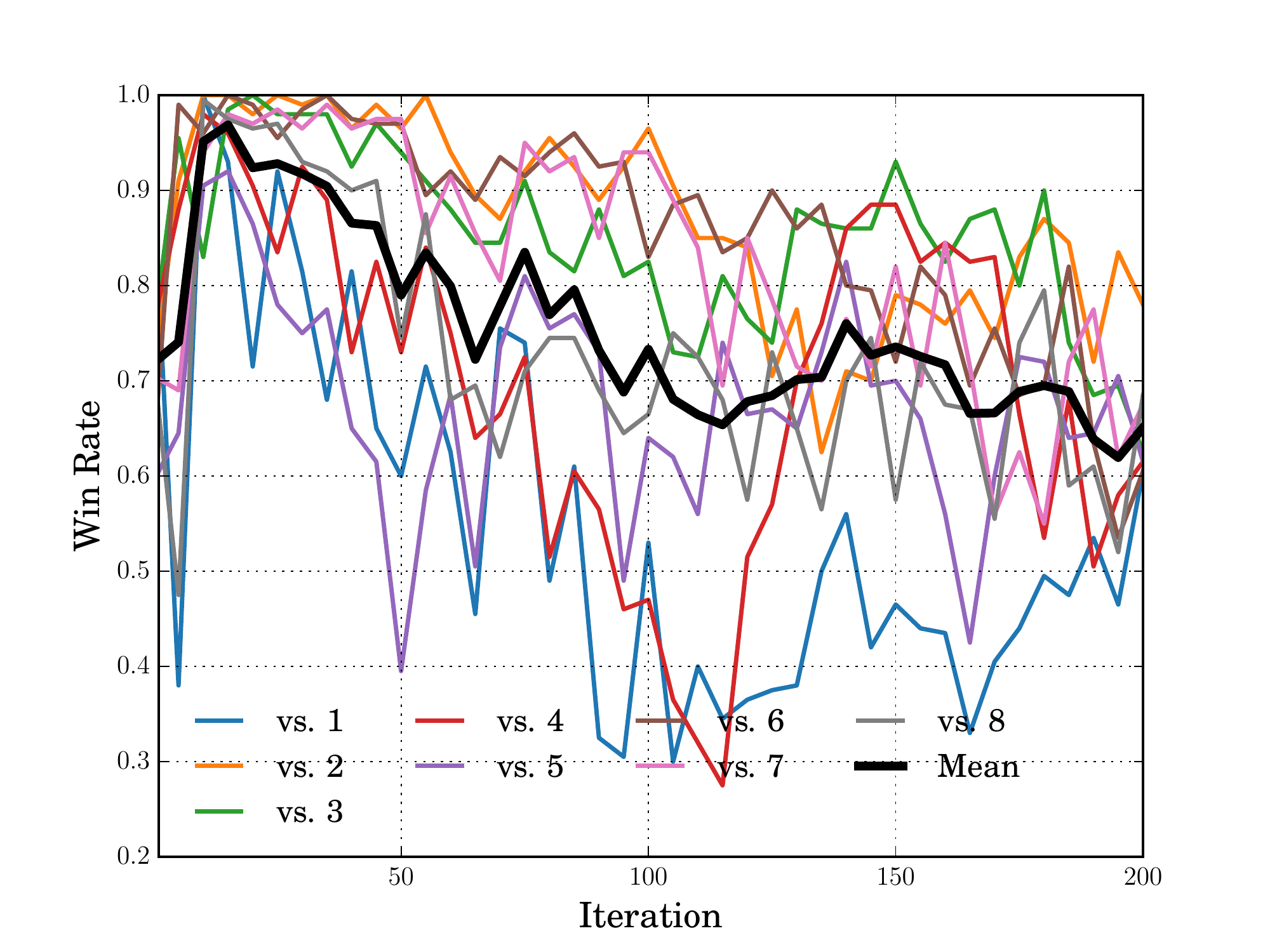}
    \caption{The win rate for PBT against all 8 baselines, this time with ML value networks instead of a single value output.}
    \label{fig:P+ML win rate}
\end{figure}

\begin{figure}[hbtp!]
    \centering
    \includegraphics[width=\columnwidth]{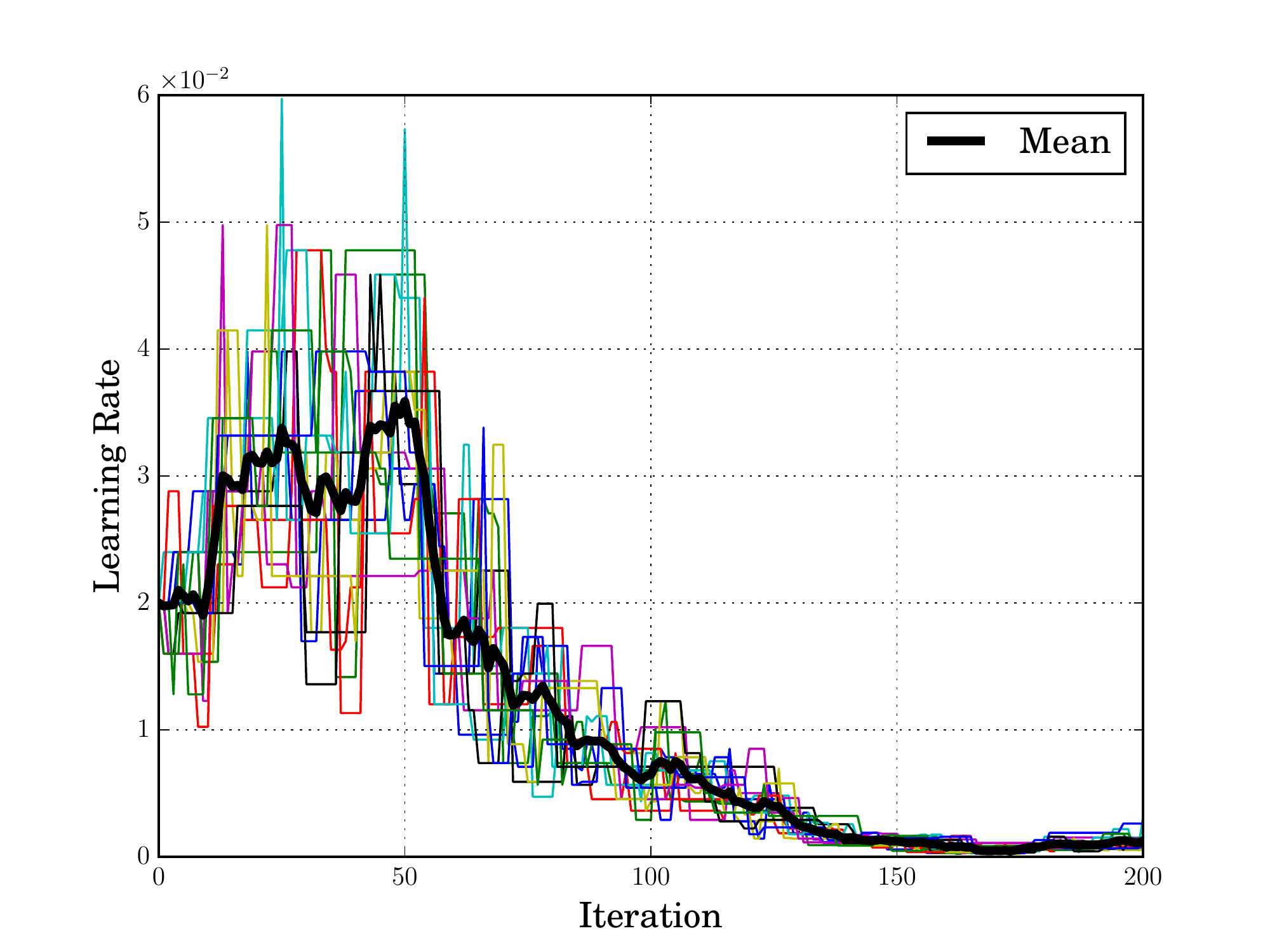}
    \caption{The learning rate curve during training for ML value networks.}
    \label{fig:P+ML learning rate}
\end{figure}

\begin{figure}[t]
    \centering
    \includegraphics[width=\columnwidth]{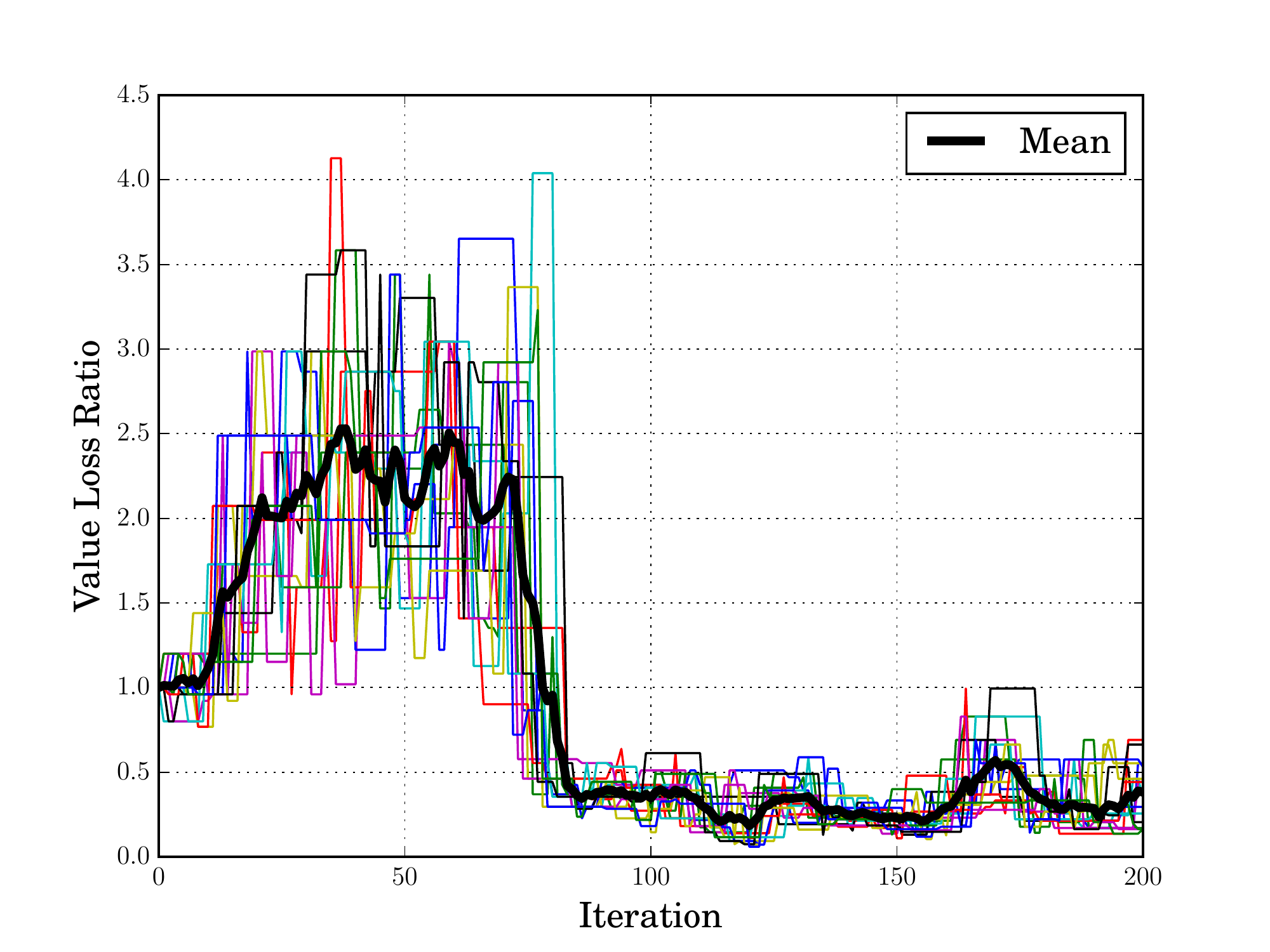}
    \caption{The value loss ratio curve during training for ML value networks.}
    \label{fig:P+ML value loss}
\end{figure}

The win rate of the PBT version against the 8 baselines, where all are trained with ML value networks, is shown in Figure \ref{fig:P+ML win rate}. The result seems similar to the previous PBT result in the main text, Figure 3, except where PBT loses against agent 1 (the best performing baseline agent) during around iterations 100 to 200. However, near the end of training, PBT performs better than all 8 baseline versions. In fact, we also evaluate this PBT version against the 16 baselines, 8 for single value (described in the section of Experiments for 9x9 Go in the main text) and 8 for MLVN (in this supplementary). Similar to Figure \ref{fig:P+ML win rate}, near the end of training, PBT still performs comparable or better than the 16 baseline versions.

Figure \ref{fig:P+ML learning rate} shows the learning rate curve. The learning rate starts from 0.02 and increases to 0.35 at iteration 50. After iteration 50, the learning rate decreases gradually for the remainder of training. Figure \ref{fig:P+ML value loss} shows the value loss ratio curve. Similarly, the value loss increases in the beginning and decreases in the middle of training. The value loss ratio stabilizes around 0.3 to 0.5 after 100 iterations. Since we train 11 labels for the ML value, the real loss ratio between the policy and the value loss should correspond to about 1:3.3 to 1:5.5, which is also close to the previous result. 
The results show that PBT can adjust to the suitable hyperparameters from different initial settings.

\bibliographystyle{aaai} \bibliography{AAAI-WuT.8297.bib}

\begin{thebibliography}{}

\bibitem[\protect\citeauthoryear{Jaderberg \bgroup et al\mbox.\egroup
  }{2017}]{Jaderberg2017PopulationBT}
Jaderberg, M.; Dalibard, V.; Osindero, S.; Czarnecki, W.; Donahue, J.; Razavi,
  A.; Vinyals, O.; Green, T.; Dunning, I.; Simonyan, K.; Fernando, C.; and
  Kavukcuoglu, K.
\newblock 2017.
\newblock Population based training of neural networks.
\newblock {\em ArXiv} abs/1711.09846.

\bibitem[\protect\citeauthoryear{Jaderberg \bgroup et al\mbox.\egroup
  }{2019}]{jaderberg2019human}
Jaderberg, M.; Czarnecki, W.~M.; Dunning, I.; Marris, L.; Lever, G.; Castaneda,
  A.~G.; Beattie, C.; Rabinowitz, N.~C.; Morcos, A.~S.; Ruderman, A.; et~al.
\newblock 2019.
\newblock Human-level performance in 3{D} multiplayer games with
  population-based reinforcement learning.
\newblock {\em Science} 364(6443):859--865.

\bibitem[\protect\citeauthoryear{Lanctot \bgroup et al\mbox.\egroup
  }{2019}]{lanctot2019openspiel}
Lanctot, M.; Lockhart, E.; Lespiau, J.-B.; Zambaldi, V.; Upadhyay, S.;
  P{\'e}rolat, J.; Srinivasan, S.; Timbers, F.; Tuyls, K.; Omidshafiei, S.;
  et~al.
\newblock 2019.
\newblock Open{S}piel: A framework for reinforcement learning in games.
\newblock {\em arXiv preprint arXiv:1908.09453}.

\bibitem[\protect\citeauthoryear{Lee \bgroup et al\mbox.\egroup
  }{2019}]{Lee2019MinigoAC}
Lee, B.; Jackson, A.; Madams, T.; Troisi, S.; and Jones, D.
\newblock 2019.
\newblock Minigo: A case study in reproducing reinforcement learning research.
\newblock In {\em RML@ICLR}.

\bibitem[\protect\citeauthoryear{Pascutto}{2017}]{Leela-zero}
Pascutto, G.-C.
\newblock 2017.
\newblock Leela-zero.
\newblock \url{https://github.com/gcp/leela-zero}.

\bibitem[\protect\citeauthoryear{Rosin}{2011}]{rosin2011multi}
Rosin, C.~D.
\newblock 2011.
\newblock Multi-armed bandits with episode context.
\newblock {\em Annals of Mathematics and Artificial Intelligence}
  61(3):203--230.

\bibitem[\protect\citeauthoryear{Silver \bgroup et al\mbox.\egroup
  }{2016}]{silver2016mastering}
Silver, D.; Huang, A.; Maddison, C.~J.; Guez, A.; Sifre, L.; Van Den~Driessche,
  G.; Schrittwieser, J.; Antonoglou, I.; Panneershelvam, V.; Lanctot, M.;
  et~al.
\newblock 2016.
\newblock Mastering the game of {G}o with deep neural networks and tree search.
\newblock {\em nature} 529(7587):484--489.

\bibitem[\protect\citeauthoryear{Silver \bgroup et al\mbox.\egroup
  }{2017}]{silver2017mastering}
Silver, D.; Schrittwieser, J.; Simonyan, K.; Antonoglou, I.; Huang, A.; Guez,
  A.; Hubert, T.; Baker, L.; Lai, M.; Bolton, A.; et~al.
\newblock 2017.
\newblock Mastering the game of {G}o without human knowledge.
\newblock {\em Nature} 550(7676):354.

\bibitem[\protect\citeauthoryear{Silver \bgroup et al\mbox.\egroup
  }{2018}]{silver2018general}
Silver, D.; Hubert, T.; Schrittwieser, J.; Antonoglou, I.; Lai, M.; Guez, A.;
  Lanctot, M.; Sifre, L.; Kumaran, D.; Graepel, T.; et~al.
\newblock 2018.
\newblock A general reinforcement learning algorithm that masters chess, shogi,
  and {G}o through self-play.
\newblock {\em Science} 362(6419):1140--1144.

\bibitem[\protect\citeauthoryear{Tian and Zhu}{2015}]{tian2015better}
Tian, Y., and Zhu, Y.
\newblock 2015.
\newblock Better computer go player with neural network and long-term
  prediction.
\newblock {\em arXiv preprint arXiv:1511.06410}.

\bibitem[\protect\citeauthoryear{Tian \bgroup et al\mbox.\egroup
  }{2019}]{Tian2019ELFOA}
Tian, Y.; Ma, J.; Gong, Q.; Sengupta, S.; Chen, Z.; Pinkerton, J.; and Zitnick,
  C.~L.
\newblock 2019.
\newblock {ELF} {O}pen{G}o: an analysis and open reimplementation of alphazero.
\newblock In {\em ICML}.

\bibitem[\protect\citeauthoryear{Wang \bgroup et al\mbox.\egroup
  }{2019}]{wang2019hyper}
Wang, H.; Emmerich, M.; Preuss, M.; and Plaat, A.
\newblock 2019.
\newblock Hyper-parameter sweep on alphazero general.
\newblock {\em arXiv preprint arXiv:1903.08129}.

\bibitem[\protect\citeauthoryear{Wu \bgroup et al\mbox.\egroup
  }{2018}]{wu2018multilabeled}
Wu, T.-R.; Wu, I.-C.; Chen, G.-W.; Wei, T.-h.; Wu, H.-C.; Lai, T.-Y.; and Lan,
  L.-C.
\newblock 2018.
\newblock Multilabeled value networks for computer {G}o.
\newblock {\em IEEE Transactions on Games} 10(4):378--389.

\bibitem[\protect\citeauthoryear{Wu}{2019}]{Wu2019AcceleratingSL}
Wu, D.~J.
\newblock 2019.
\newblock Accelerating self-play learning in {G}o.
\newblock {\em ArXiv} abs/1902.10565.

\end{thebibliography}

\end{document}